\documentclass[runningheads]{llncs}
\usepackage[T1]{fontenc}
\usepackage{graphicx}

\usepackage{amsmath,amssymb,mathtools}

\usepackage{booktabs}
\usepackage{enumitem}

\usepackage[caption=false]{subfig} 
\usepackage{algorithm}
\usepackage{algpseudocode}
\usepackage{tabularx} 
\newcolumntype{Y}{>{\centering\arraybackslash}X}

\title{Training-Free Multi-Style Fusion Through Reference-Based Adaptive Modulation}
\titlerunning{Training-Free Multi-Style Fusion}
\author{Xu Liu\inst{1}\thanks{Corresponding author.},
Yibo Lu\inst{2}, Xinxian Wang\inst{3},
Xinyu Wu\inst{1}}
\authorrunning{Xu Liu et al.}
\institute{
University of Washington, Seattle, WA 98195, USA\\
\email{\{xliu28,xinyuw23\}@uw.edu}
\and 
Nanjing University of Posts and Telecommunications, Jiangsu, Nanjing, China\\
\email{p23000309@njupt.edu.cn}
\and
Nanjing Foreign Language School, Jiangsu, Nanjing, China\\
\email{wxx2027f@163.com}
}
\begin{document}  
\pagestyle{empty}
\maketitle
\thispagestyle{empty}
\begin{abstract}
We propose Adaptive Multi\-Style Fusion (AMSF), a reference\-based training-free framework that enables controllable fusion of multiple reference styles in diffusion models. Most of the existing reference-based methods are limited by
(a) acceptance of only one style image, thus prohibiting hybrid aesthetics and scalability to more styles, and
(b) lack of a principled mechanism to balance several stylistic influences. 
AMSF mitigates these challenges by encoding all style images and textual hints with a semantic token decomposition module that is adaptively injected into every cross-attention layer of an frozen diffusion model.
A similarity-aware re-weighting module then recalibrates, at each denoising step, the attention allocated to every style component, yielding balanced and user-controllable blends without any fine-tuning or external adapters.
Both qualitative and quantitative evaluations show that AMSF produces multi-style fusion results that consistently outperform the state-of-the-art approaches, while its fusion design scales seamlessly to two or more styles.
These capabilities position AMSF as a practical step toward expressive multi-style generation in diffusion models.
\keywords{Multi-style fusion \and Diffusion models \and Training-free modulation}
\end{abstract}

\section{Introduction}

Text-to-image (T2I) generative models have rapidly advanced in their ability to synthesize photorealistic or stylized images from textual prompts. Building on diffusion-based architectures~\cite{ho2020denoising,rombach2022high,rout2024beyond,peebles2023scalable,yan2024diffusion,saharia2022photorealistic,ramesh2022hierarchical}
, recent methods have introduced the use of style reference images to guide the visual appearance of generated outputs~\cite{hertz2024style,rout2024rb,sohn2023styledrop,ye2023ip}. These models enable consistent style reproduction by injecting image-level features or style embeddings into the generation pipeline, a concept that also finds parallels in a wide range of visual understanding models~\cite{chen2018adversarial,hu2018squeeze,wang2025seal}.

\begin{figure}[!t]
  \centering
  \includegraphics[width=0.85\linewidth]{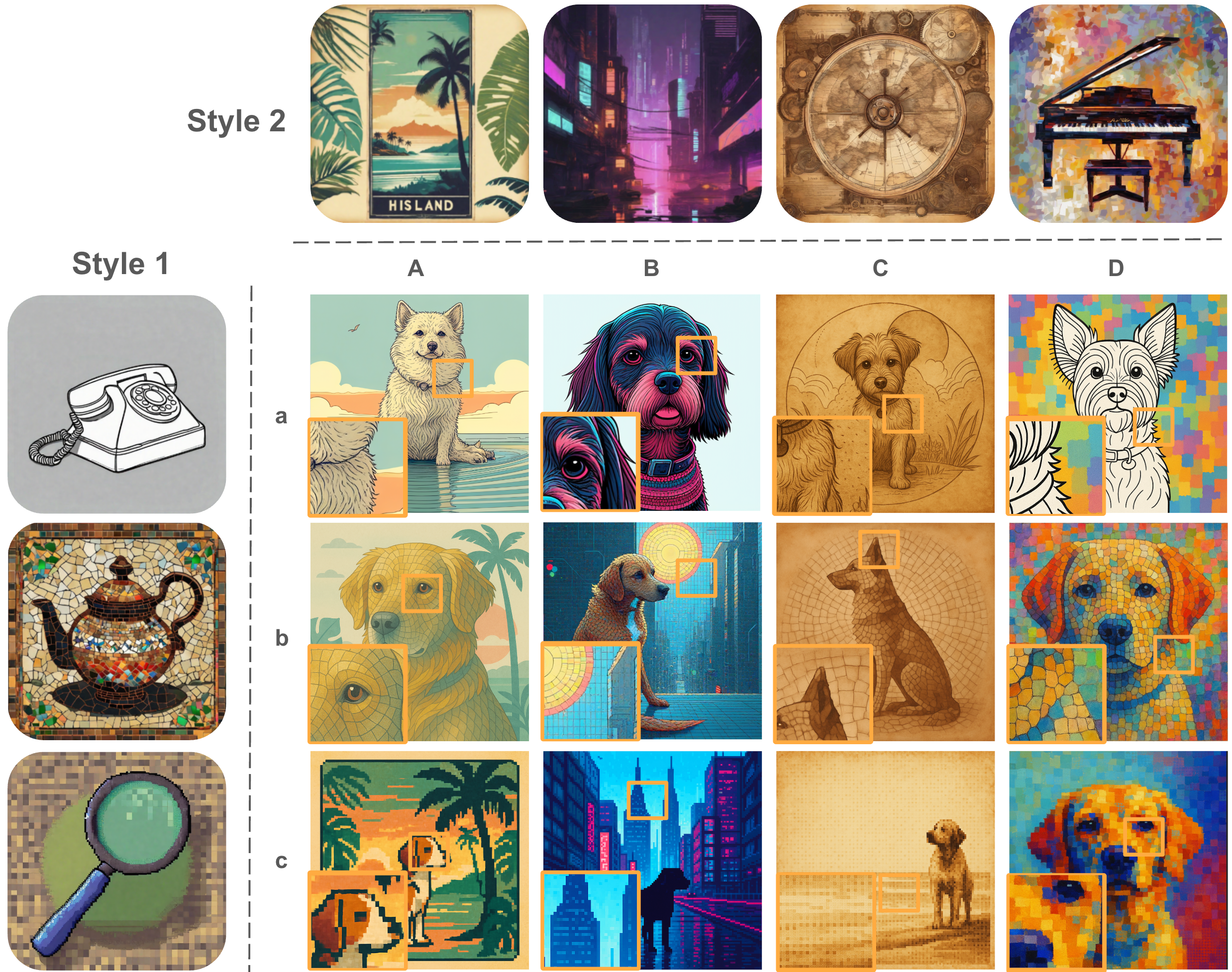}
  \caption{Overall qualitative results of the proposed Adaptive Multi-Style Fusion (AMSF) for two-style fusion.}
  \label{fig:qualattive_res}
\end{figure}

While single-style conditioning methods have shown effectiveness, they fall short in scenarios where creators seek to blend multiple distinct styles. In practice, artists and designers often wish to explore hybrid aesthetics. For example, combining the color palette of an old poster with the abstraction of a cartoon sketch, or fusing 3D renderings with watercolor textures. Current frameworks offer little support for such multi-style synthesis, despite rising demand driven by the need for greater creative flexibility, deeper personalization, and finer controllability. Users increasingly desire the ability to express complex visual ideas that do not conform to a single stylistic identity.

Moreover, controllability is essential for iterative and purposeful design processes, where the influence of each style or content prompt must be modulated in a transparent and flexible manner.

Most current T2I models with style conditioning~\cite{hertz2024style,rout2024rb,sohn2023styledrop,ye2023ip} support only one reference image and treat style as a single fixed attribute throughout the generation process. These models are unable to capture the nuanced combinations that arise when multiple stylistic elements must be harmonized. The generated outputs often overfit to a dominant reference, resulting in unbalanced or poorly blended compositions. Furthermore, existing models offer limited user control over how each reference influences the final image, making them less suitable for creative or exploratory applications where user intent varies across iterations.

To address these challenges, we propose a novel framework called Adaptive Multi-Style Fusion (AMSF). We propose to extend conventional T2I pipelines by conditioning on semantically decomposed components in a reference-based manner, followed by a Similarity Attention Reweighting (SAR) module to flexibly assign attention weights to each component. AMSF dynamically computes the influence of each reference during generation, while minimizing the disentanglement of different references (e.g., styles and subject). By integrating both visual and textual styles in an adaptive and interpretable manner, our model produces outputs that are both stylistically rich and semantically coherent. Besides, AMSF is extensible and can support more than two style references (Fig. \ref{fig:three_style}), enabling users to experiment with complex stylistic combinations in highly personalized image synthesis.

Figure~\ref{fig:qualattive_res} demonstrates the strength of our \textbf{Adaptive Multi-Style Fusion (AMSF)} across different style compositions. All generated with the fixed subject prompt \emph{“a dog”}. AMSF consistently blends color palettes, stroke geometry, and textures into unified, visually pleasing compositions.
In the upper-right (aD) example, AMSF fuses an abstract art background with a line drawing subject: the dog preserves crisp line-drawing contours while the backdrop retains expressive abstract brushstrokes. In the lower-left (cA) example, pixel style block geometry remains evident, while the warm palette and stylized coconut trees clearly evoke a vintage travel poster aesthetic.
These examples show that AMSF balances each reference throughout the diffusion trajectory, preserving the signature traits of every contributing style.

In summary, our key contributions are:

\begin{enumerate}
    \item We present a novel reference-based multi-style fusion method for image stylization, which adaptively fuses information from multiple visual and textual style references to produce a coherent embeddings that preserves details of each style.
    \item We introduce an efficient Similarity-aware Attention Re-weighting (SAR) module that eliminates the need for manual hyperparameter tuning. This innovation leads to superior style coherence and content fidelity in the multi-style generation task.
    \item We perform extensive experiments and user studies that cover both style/prompt alignment and overall aesthetic quality, demonstrating AMSF's state-of-the-art performance in multi-style generation.

\end{enumerate}
\section{Related Work}

\subsection{Training-free Stylization}
Training-free stylization avoids fine-tuning and favors fast deployment. RF-Inversion~\cite{rout2024semantic} performs semantic inversion/editing via rectified flows. FlowEdit~\cite{kulikov2024flowedit} constructs ODE transports using pre-trained flows. FreeTuner~\cite{xu2024freetuner} separates subject structure rendering and style injection via intermediate attention, enabling zero-shot multi-style editing.

FreeTuner~\cite{xu2024freetuner} disentangles generation into a two-stage process—first rendering subject structure from text prompts, then injecting style guidance via intermediate diffusion attention. This enables zero-shot, multi-subject and multi-style image editing without any fine-tuning.

\subsection{Training-based Stylization}
Training-based methods enhance stylization quality through lightweight adapters or task-specific fine-tuning. IP-Adapter~\cite{ye2023ip} introduces a light-weighted adapter with decoupled image and text cross-attention, enabling image-prompt control without altering the diffusion backbone. InstantStyle~\cite{wang2024instantstyle} is a tuning‑free style personalization framework that decouples content and style via CLIP‑based~\cite{radford2021learning,qiu2021vt} feature subtraction and injects image features only into empirically identified “style blocks” of the U‑Net~\cite{ronneberger2015u}
. This achieves effective style transfer while preserving text prompt alignment and avoiding content leakage. DEADiff~\cite{qi2024deadiff} trains Q-Former modules on paired data to disentangle content and style, injecting them into distinct attention paths for more precise style control under textual guidance. U-StyDiT~\cite{zhang2025u} is built on a DiT-based transformer diffusion backbone and employs a Multi-view Style Modulator (MSM) to extract style cues from both global and local image patches. It introduces a StyDiT Block, which integrates style, content, and Canny tokens via multi-modal attention to disentangle structure and style. The model is jointly fine-tuned—alongside U-Net and the text encoder—on a large-scale Aes4M dataset, enabling control over global tone and fine-grained textures for ultra-high quality artistic stylization.

\subsection{Multi-Style Fusion}
MSG-Net~\cite{zhang2017msgnet} aligns second-order statistics for real-time transfer. Multi-LoRA~\cite{zhong2024multi} composes pre-trained LoRAs without retraining. Style Mixer~\cite{huang2019style} uses multi-level fusion and patch-attention with region-wise mixing. Unlike these, AMSF separates content/style streams and dynamically balances arbitrary numbers of styles without training.

Our method addresses these challenges by separating content and style streams and applying similarity-aware dynamic fusion, enabling seamless, arbitrary multi-style stylization without retraining.

\section{Preliminaries}

\subsection{Diffusion Models}

Diffusion models generate images by reversing a gradual noising process, typically modeled as a Stochastic Differential Equation (SDE)~\cite{song2020score}. The forward process is defined as:

\begin{equation}
dX_t = f(X_t, t) dt + g(X_t, t) dW_t, \quad X_0 \sim p_0,
\end{equation}

Stable Diffusion~\cite{rombach2022high} is a latent diffusion model that operates in a compressed latent space, enabling efficient high-resolution image synthesis. It conditions generation on text embeddings via cross-attention mechanisms. StableCascade\cite{pernias2023wurstchen} extends this approach with a three-stage pipeline, achieving higher compression rates and faster generation while maintaining image quality.

\subsection{Reference-Based Modulation}

Recent training-free reference-based modulation~\cite{rout2024rb} reframes sampling from a pre-trained diffusion model as a stochastic optimal-control (SOC) problem with style feature merging.  

An attention feature aggregation (AFA) block is inserted in each cross-attention layer: keys/values from text, style, and optionally a content image are processed in parallel and averaged, preventing content leakage from the style image while preserving prompt guidance.
RB-Modulation therefore delivers high-fidelity, style-faithful images with zero fine-tuning, but it is intrinsically single-reference—balancing multiple, potentially conflicting styles is not addressed.  Our AMSF framework retains the SOC–AFA philosophy yet scales it to an arbitrary set of style references.

\section{Method}

We first give an overview of the proposed
Adaptive Multi-Style Fusion (AMSF)
in Section~\ref{sec:overview}.
The method contains two modules:
Semantic Decomposition and
Similarity-aware Attention Re-weighting (SAR),
presented in Sections~\ref{sec:moduleA} and~\ref{sec:moduleB}.


\begin{figure}[htbp]
  \centering
  \includegraphics[width=\linewidth]{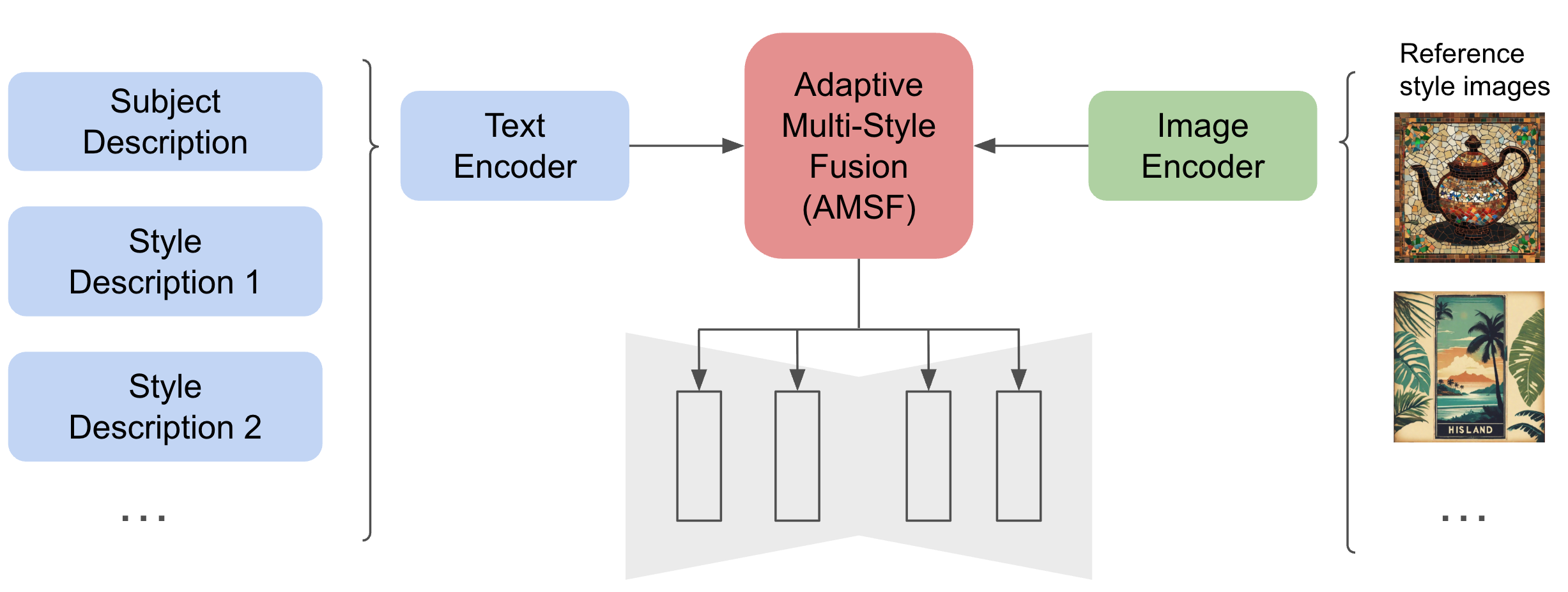}
  \caption{The overall architecture of the proposed Adaptive Multi-Style Fusion (AMSF). Subject and style descriptions are decomposed and processed by a text encoder separately, while reference images are processed by an image encoder. All the embeddings are adaptively merged before updating the attention features in the denoising network.}
  \label{fig:msh_cf_big_overview}
\end{figure}

\subsection{Overview}
\label{sec:overview}

AMSF extends a text-to-image diffusion backbone so that a user can
blend several style references in one pass, see Figure~\ref{fig:msh_cf_big_overview}.
AMSF takes reference style images, textual style descriptions, and one subject prompt as the input.
All resulting tokens are merged once and adaptively reused
by every cross-attention layer, allowing the relative strength of each
reference to be re-weighted on the fly without re-encoding.

\subsection{Semantic Token Decomposition}
\label{sec:moduleA}

\begin{figure}[t]
  \centering
  \includegraphics[width=1\linewidth]{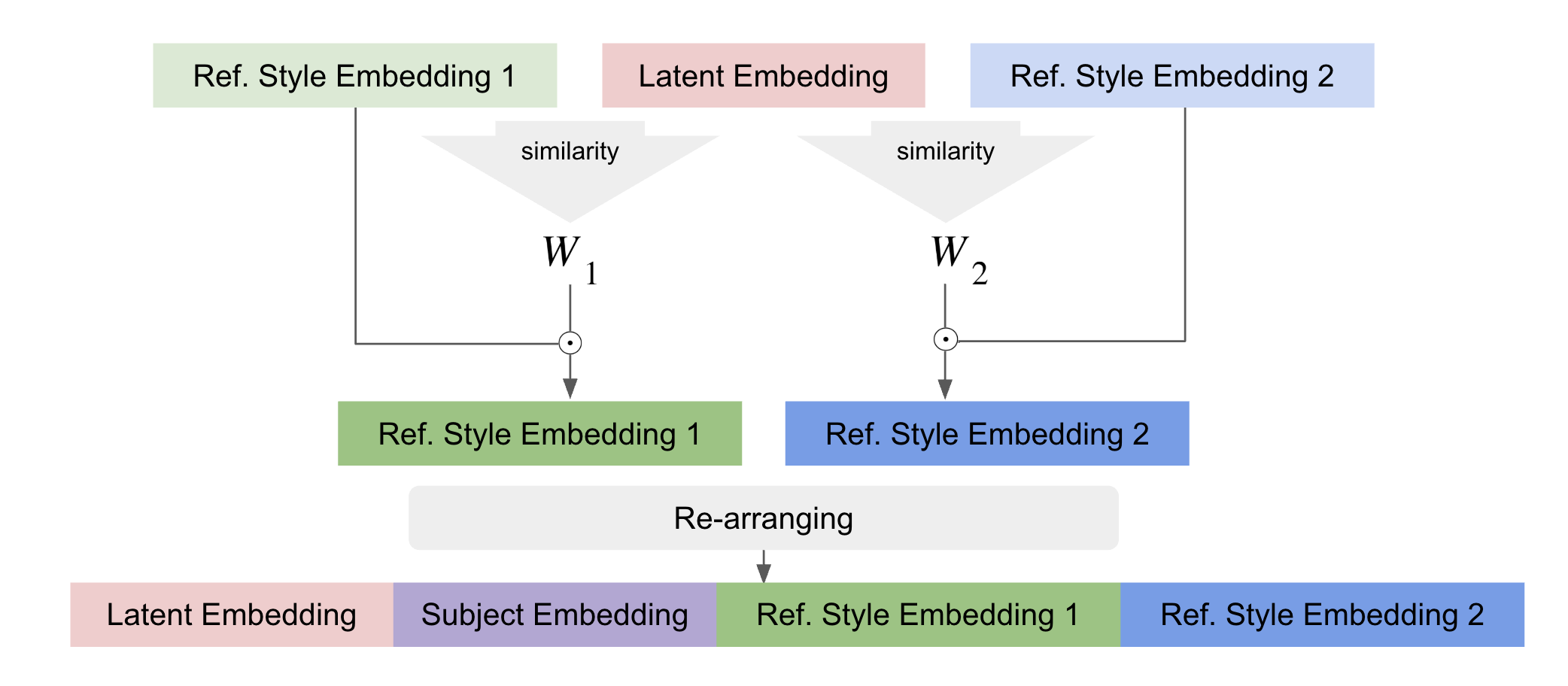}
  \caption{Illustration of the Similarity-aware Attention Re-weighting (SAR) module. The SAR module adaptively re-weights reference style embeddings based on their similarity to a latent embedding. These re-weighted style embeddings are then rearranged and concatenated with the latent and subject (e.g., “dog”) embeddings for use in attention layers. For simplicity, style embedding contains both text and image embedding, and we only demonstrate the case for two styles.}
  \label{fig:msh_cf_overview}
\end{figure}

For single-style transfer, the input normally consists of a style prompt
and a reference image.  
A naive extension to two-style fusion would be concatenating different style tokens together.
For example, in Figure~\ref{fig:prompt_fail}, the repeated subject token ``dog'' receives twice the
attention as it appears in both prompts, overwhelming the style tokens and impairing stylization.
We therefore decompose the inputs into five distinct components. Specifically, let
\(I_{1},I_{2}\) be the two style images,
\(T_{1},T_{2}\) their corresponding style prompts (e.g., ``mosaic style'' ),
and \(T_{s}\) the subject prompt (e.g., ``dog'' ).
We first extract corresponding embeddings using the image encoder $\mathcal{E}_I$ and text encoder $\mathcal{E}_T$. 
All the extracted tokens are stacked in the fixed order in the latent representation (Fig.~\ref{fig:msh_cf_overview})

\begin{equation}
  \mathbf{Z}= \bigl[\mathcal{E}_T(T_{1});\mathcal{E}_T(T_{2});\mathcal{E}_T(T_{s});\mathcal{E}_I(I_{1});\mathcal{E}_I(I_{2})\bigr]
\end{equation}

This decomposition provides better style-subject balance, since the subject prompt is isolated in $\mathcal{E}_T(T_{s})$, preventing its weight from being unintentionally doubled. Consequently, Our AMSF achieves visually balanced hybrids where details of multiple styles can be effectively captured while maintaining subject prompt alignment.

\subsection{Similarity-aware Attention Re-weighting}
\label{sec:moduleB}

While $\mathbf{Z}$ carries all references, we still need to decide—at
each denoising step—the weight to be assigned to each style.
We propose Similarity-aware Attention Re-weighting (SAR) inspired by \cite{hu2018squeeze,woo2018cbam} in Figure~\ref{fig:msh_cf_overview} to solve
this problem by measuring how similar the current latent features are to
each style token and then assigning weights accordingly.

Let $\mathbf{x}\in\mathbb{R}^{HW\times C}$ be the query features in a
cross-attention layer, where $H$ and $W$ are spatial height and width
and $C$ is the channel dimension. $\mathbf{x}_{j}\in\mathbb{R}^{C}$ for the feature vector at spatial
index $j$ ($j=1,\dots,HW$).
Denote by sim, the cosine similarity,
by $\bar{\mathbf{x}}$ the spatial mean of~$\mathbf{x}$,
and by $\mathbf{s}_{i}$ the style token of the
$i$-th reference style.
\begin{equation}
  \sigma_{i}= \operatorname{sim}\bigl(\bar{\mathbf{x}},\mathbf{s}_{i}\bigr),\quad
  \tau_{i}= \frac{1}{HW}\sum_{j}\operatorname{sim}\bigl(\mathbf{x}_{j},
                                                  \mathbf{s}_{i}\bigr)
\end{equation}
The pair $(\sigma_i,\tau_i)$ therefore captures \emph{global} and
\emph{token-level} agreement between the current latent and style~$i$.

Taken two style fusion as an example, our initial design split the cross-attention weights evenly among the
three components in the latents—subject, style 1, and style 2—by fixing all weights to
$33.3\,\%$:
\begin{equation}
w_{\text{subject}}=w_{1}=w_{2}=\tfrac13.
\end{equation}

We observed that a dominant reference (e.g., \emph{mosaic}) can quickly
overwhelm a weaker one.  We therefore define an \emph{adaptive} damping
term:
\begin{equation}
\gamma_{\text{auto}}
  = \operatorname{clamp}\!\bigl(
      1 \;+\; \kappa\,\lvert\sigma_{1}-\sigma_{2}\rvert
        +      \lvert\tau_{1}-\tau_{2}\rvert,\,
      \gamma_{\min},\gamma_{\max}\bigr),
\label{eq:autogamma}
\end{equation}
with lower/upper guards $\gamma_{\min},\gamma_{\max}$.

\smallskip
The leading “\(1\)” gives every step a baseline penalty so
$\gamma_{\text{auto}}\!\ge\!1$ even when the two styles are identical.
The factor $\kappa$ simply balances the more reliable global
gap \(\lvert\sigma_{1}-\sigma_{2}\rvert\) against the noisier
token-level gap \(\lvert\tau_{1}-\tau_{2}\rvert\); the value was chosen
by a coarse grid-search on the validation set.

When one style begins to dominate, the discrepancy terms increase,
$\gamma_{\text{auto}}$ rises toward $\gamma_{\max}$, and the stronger style should be down-weighted, allowing the suppressed style to
re-emerge. Therefore, each style $i$ receives a similarity score:

\begin{equation}
\text{score}_{i}=
\frac{(1+\sigma_{i})(1+\tau_{i})}{
      1+\lVert\mathbf{s}_{i}\rVert^{\,\gamma_{\text{auto}}}},
\end{equation}

\noindent and its final attention weight is the normalised ratio

\begin{equation}
w_{i}=\frac{\text{score}_{i}}
           {\text{score}_{1}+\text{score}_{2}+\delta},
\end{equation}

where $\delta$ is a small constant that avoids division by zero.
If both styles match the current latent equally well,
$w_{1}\!\approx\!w_{2}$.
If one style overwhelms the other, the increased $\gamma_{\text{auto}}$ lowers that
style’s score, restoring balance.  
Empirically, this automatic scheme is far more reliable and efficient than fixed or
hand-tuned weights, particularly when the two references differ greatly
in color, texture, or abstraction level.

\begin{figure}[t!]
  \centering
  \includegraphics[width=\linewidth]{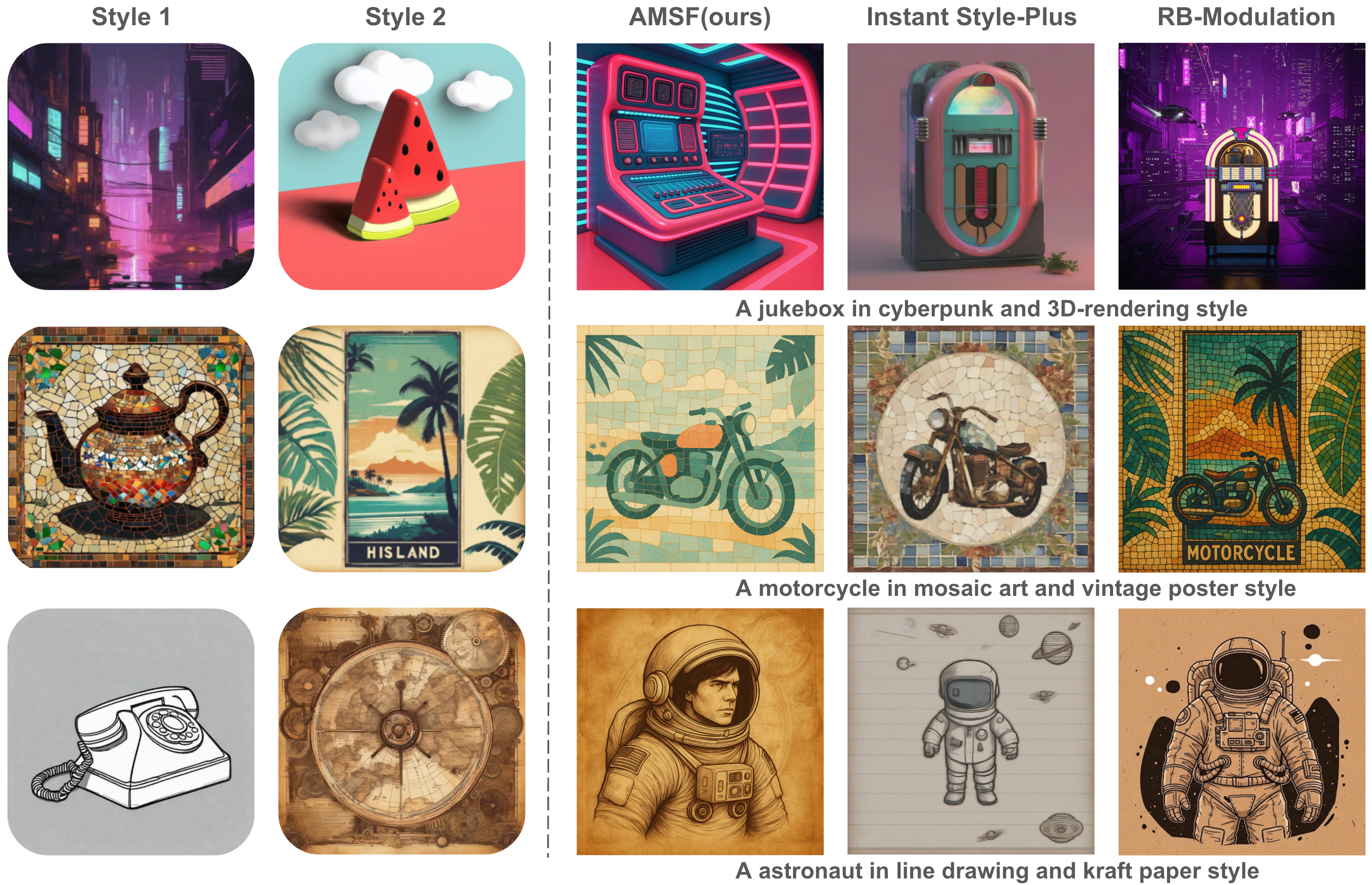}
  \caption{Comparison between AMSF and baseline methods.}
  \label{fig:baseline_test}
\end{figure}

\section{Experiment}
\label{exp}
\subsection{Baselines and dataset}
\noindent \textbf{Baselines: }
We compare our approach with state-of-the-art diffusion-based image stylization methods, including RB\-Modulation~\cite{rout2024rb}, DEADiff~\cite{qi2024deadiff}, InstantStyle\-Plus~\cite{wang2024instantstyle}, and IP\-Adapter~\cite{ye2023ip}. We focus primarily on 2-style composition, with additional multi-style results presented in the ablation study.
For baselines that are designed for single-style transfer, we adapt them by linearly interpolating the embeddings of two different style images in the feature space. Specifically, the image embeddings of the two input styles are averaged before applying model processing. We include both style descriptions in the prompt for all methods, formatted as: ``in [description\_1] style and [description\_2] style.''

\noindent \textbf{Dataset: }
We use the StyleAligned dataset~\cite{hertz2024style}, which contains 100 style images paired with textual descriptions. Since not all style combinations are semantically meaningful (e.g., mixing two 2D styles with conflicting color schemes), we selected 50 coherent style pairs for evaluation. Each style pair is tested across 100 different prompts randomly drawn from the dataset, resulting in a total of 5,000 outputs for quantitative comparison.

\noindent\textbf{Metrics:}
Following \cite{rout2024rb}, we use CLIP-T~\cite{kwon2022clipstyler} to measure prompt alignment, and DINO~\cite{caron2021emerging} for style alignment of each reference style. Besides, to quantify the balance between the two styles, we calculate the harmonic mean (HM) of the CLIP-T or DINO scores. A small HM score suggests an imbalance, where one reference style predominantly influences the output.

\begin{table}[t]
  \centering
  \small
  \setlength{\tabcolsep}{3pt}
  \caption{Quantitative results of two-style fusion. For each style, we report prompt alignment (CLIP-T) and style similarity (DINO). The Harmonic Mean (HM) assesses balance; lower HM indicates dominance by one style.}
  \label{tab:two-style-metrics}
  \begin{tabularx}{0.98\textwidth}{@{}l*{3}{Y}*{3}{Y}@{}}
    \toprule
    & \multicolumn{3}{c}{\textbf{CLIP-T score}} & \multicolumn{3}{c}{\textbf{DINO score}} \\
    \cmidrule(lr){2-4}\cmidrule(l){5-7}
    & style1 & style2 & HM & style1 & style2 & HM \\
    \midrule
    \textbf{IP-Adapter}         & 0.19 & 0.15 & 0.17 & 0.69 & 0.61 & 0.63 \\
    \textbf{InstantStyle-Plus}  & 0.21 & 0.18 & 0.19 & 0.70 & 0.65 & 0.67 \\
    \textbf{DEADiff}            & 0.17 & 0.16 & 0.16 & 0.68 & 0.63 & 0.65 \\
    \textbf{RB-Modulation}      & 0.23 & 0.20 & 0.20 & 0.70 & 0.66 & 0.68 \\
    \textbf{AMSF (Ours)}        & \textbf{0.24} & \textbf{0.23} & \textbf{0.24}
                                & \textbf{0.72} & \textbf{0.73} & \textbf{0.72} \\
    \bottomrule
  \end{tabularx}
\end{table}

\noindent\textbf{Experimental settings:}\;
All experiments are performed on a single NVIDIA A100 GPU.  
For every task we keep the AMSF hyper-parameters fixed at
\(\gamma_{\min}=1\), \(\gamma_{\max}=5\), and \(\kappa=4\) in
Eq.~\eqref{eq:autogamma}.  
Unless otherwise noted, all evaluations use a two-style fusion setup;
multi-style results appear only in the ablation study
(Section~\ref{sec:ablation}).

\begin{table}[t]
  \centering
  \small
  \setlength{\tabcolsep}{3pt} 
  \caption{User study results. Participants judged Overall Quality (OQ), Style Alignment (SA), and Prompt Alignment (PA). AMSF is preferred across metrics.}
  \label{tab:user-study}
  \begin{tabularx}{\textwidth}{@{}l*{9}{Y}@{}}
    \toprule
    & \multicolumn{3}{c}{\textbf{RB-Modulation}} 
    & \multicolumn{3}{c}{\textbf{DEADiff}} 
    & \multicolumn{3}{c}{\textbf{InstantStyle-Plus}} \\
    \cmidrule(lr){2-4}\cmidrule(lr){5-7}\cmidrule(l){8-10}
    \textbf{Human Preference (\%)} 
      & \textbf{OQ}\,$\uparrow$ & \textbf{SA}\,$\uparrow$ & \textbf{PA}\,$\uparrow$
      & \textbf{OQ}\,$\uparrow$ & \textbf{SA}\,$\uparrow$ & \textbf{PA}\,$\uparrow$
      & \textbf{OQ}\,$\uparrow$ & \textbf{SA}\,$\uparrow$ & \textbf{PA}\,$\uparrow$ \\
    \midrule
    \textbf{Alternative} & 32.6 & 23.4 & 39.2 & 25.2 & 21.8 & 33.4 & 40.2 & 25.5 & 27.9 \\
    \textbf{Tie}         & 19.3 &  9.3 & 15.9 & 12.5 &  8.2 &  9.5 &  7.9 & 10.9 & 11.7 \\
    \textbf{AMSF (Ours)} & \textbf{48.1} & \textbf{67.3} & \textbf{44.9}
                         & \textbf{62.3} & \textbf{70.0} & \textbf{57.1}
                         & \textbf{51.9} & \textbf{63.6} & \textbf{60.4} \\
    \bottomrule
  \end{tabularx}
\end{table}

\subsection{Qualitative Results}

In Figure~\ref{fig:baseline_test}, we compare AMSF with the strongest baselines InstantStyle-Plus and RB-Modulation on three different prompts that demand the harmonious fusion of two divergent references.  
In the first row, AMSF renders a jukebox whose neon trims, volumetric highlights, and metallic reflections jointly convey both the cyber-punk palette and the depth cues of a 3D render; InstantStyle-Plus largely loses the neon ambience, while RB-Modulation fails to capture the 3D style.
The second row shows a motorcycle conditioned on mosaic art and vintage-poster aesthetics. AMSF integrates the tessellated texture with the muted, poster-like colourway, whereas InstantStyle-Plus result is dominated by the mosaic style
and RB-Modulation produces oversaturated poster text that overwhelms the mosaic motif.  
The third row depicts an astronaut in line-drawing and kraft-paper style. AMSF preserves the pen-and-ink contours and the warm fibre texture of kraft simultaneously; InstantStyle-Plus drifts toward a notebook sketch with faint tones, and RB-Modulation overlays line art on a flat brown background, forfeiting the organic paper grain.  
Across all cases AMSF maintains subject geometry, balances both references without style dominance, and avoids artifacts that distract from the intended composition, illustrating its qualitative superiority over other approaches.

\subsection{Quantitative Results}
\label{sec:quant}

Table~\ref{tab:two-style-metrics} highlights a significant imbalance in the performance of baseline models across different styles. While the dominant style can vary, most baselines are generally skewed towards style 1. For instance, the averaged CLIP-T shows scores of 0.23 for style 1 versus 0.20 for style 2 in RB-Modulation, and the averaged DINO score for InstantStyle-Plus is 0.70 for style 1 compared to 0.65 for style 2. This imbalance means that even if the dominant style flips between style 1 and style 2, the overall scores for the baselines remain low.
This imbalance issue negatively impacts their harmonic means, which stay below 0.21 for CLIP-T and 0.68 for DINO.
In contrast, AMSF demonstrates nearly equal performance across both styles. Its CLIP-T scores are 0.24 for style 1 and 0.23 for style 2, while its DINO scores are 0.72 for style 1 and 0.73 for style 2. This balance results in the highest harmonic means: 0.24 for CLIP-T and 0.72 for DINO. These findings confirm that AMSF effectively improves fusion results for both styles simultaneously, successfully overcoming the style-dominance effect seen in other methods.

\begingroup

\begin{figure}[t]
  \centering
  \subfloat[Naive Token Concatenation\label{fig:prompt_fail_a}]{%
    \includegraphics[width=0.44\textwidth]{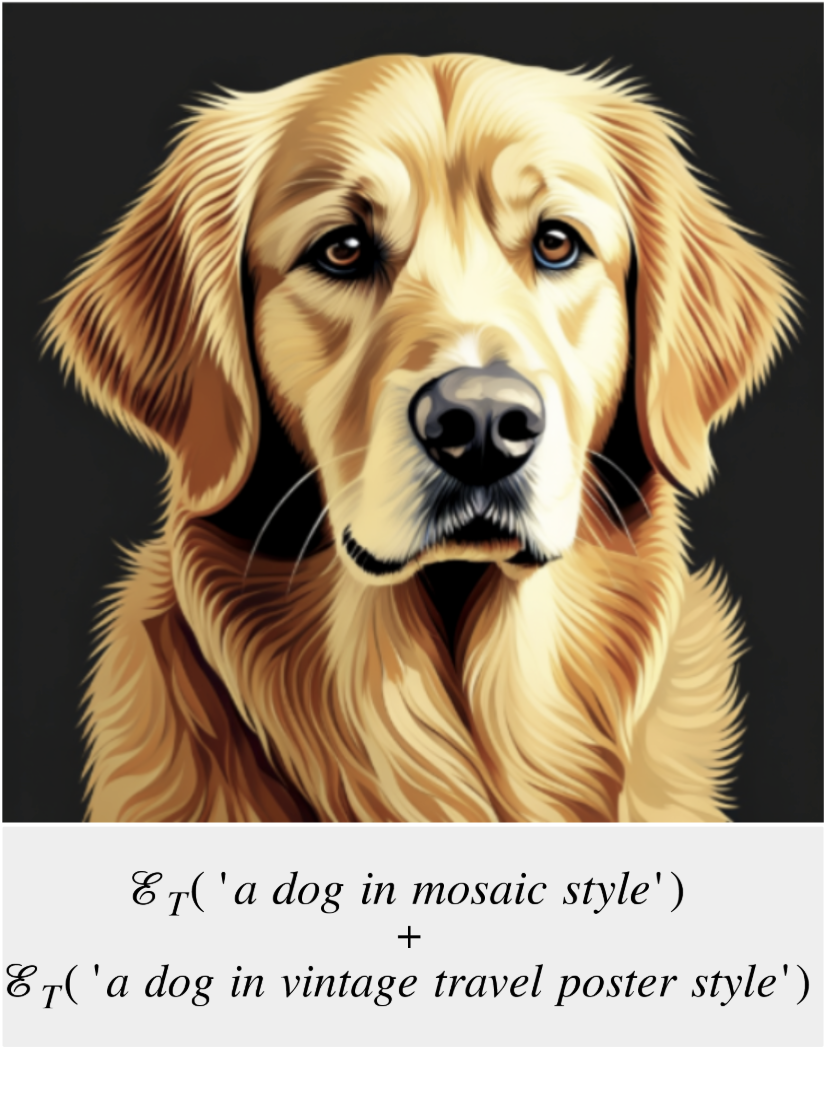}
  }\hfill
  \subfloat[Semantic Token Decomposition\label{fig:prompt_fail_b}]{%
    \includegraphics[width=0.44\textwidth]{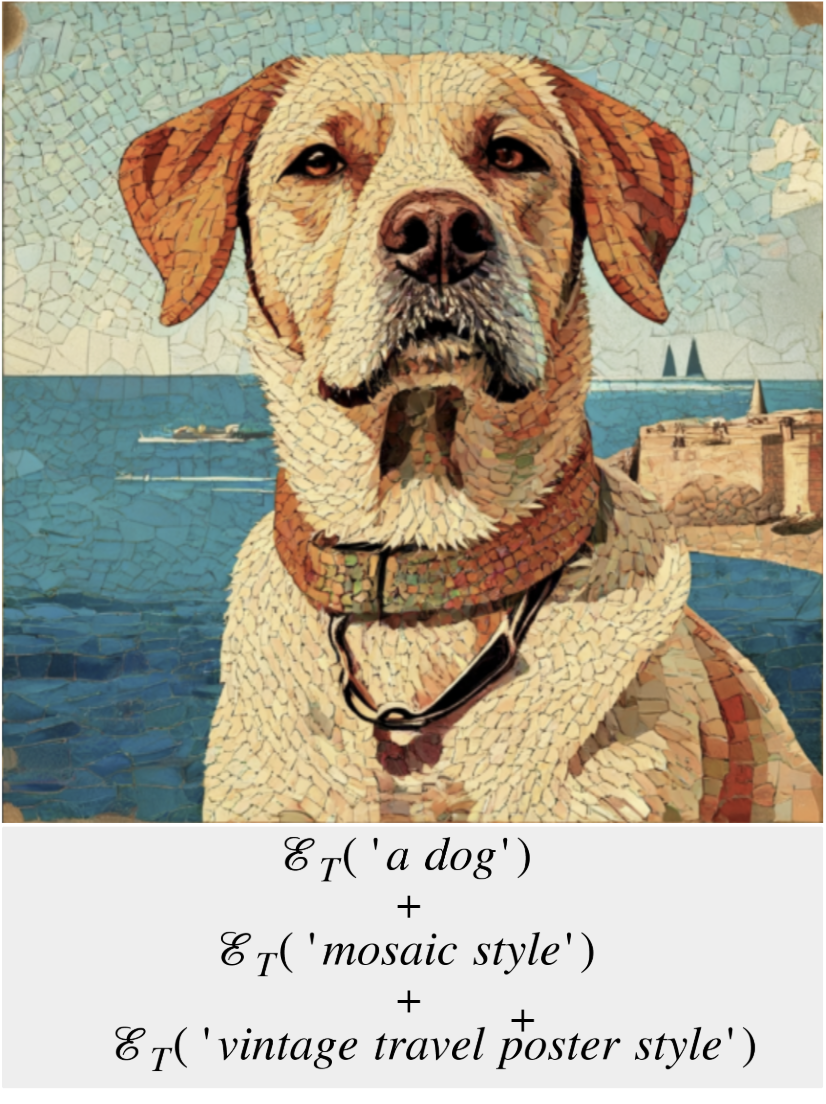}
  }
  \caption{Impact of Semantic Token Decomposition. (a) Simply concatenating the text tokens of different style prompts duplicates the subject tokens (``dog''), leading to subject dominance and loss of style. (b) Decomposing components balances subject and style appearance in the final output.}
  \label{fig:prompt_fail}
\end{figure}
\endgroup

\subsection{User Study}
To rigorously evaluate the perceptual quality of our AMSF method, we conducted a comprehensive user study, following the experimental setup described in RB-Modulation~\cite{rout2024rb}. We collected 5 responses per question, where raters were tasked with comparing AMSF-generated images directly against those produced by baseline methods. Participants were instructed to choose their preferred results based on three key criteria: overall quality (OQ), style alignment (SA), and prompt alignment (PA). This procedure yielded a total of 75,000 responses across 5,000 distinct comparisons, involving 170 participants recruited via Amazon Mechanical Turk.

The quantitative results presented in Table~\ref{tab:two-style-metrics} further confirm AMSF's superior performance. Compared to RB-Modulation, AMSF significantly excels in Style Alignment (67.3\% vs. 23.4\%) and also leads in Prompt Alignment (44.9\% vs. 39.2\%). Futhermore, AMSF received higher Overall Quality votes (48.1\% vs. 32.6\%), with ties accounting for 19.3\%. Against DEADiff~\cite{qi2024deadiff}, AMSF clearly outperforms across all criteria: securing 62.3\% of Overall Quality, 70.0\% in Style Alignment, and 57.1\% in Prompt Alignment. DEADiff's corresponding scores are substantially lower at 25.2\%, 21.8\%, and 33.4\%, respectively. Comparisons with InstantStyle-Plus~\cite{wang2024instantstyle} show AMSF again strongly preferred in Style Alignment (63.6\% vs. 25.5\%) and Prompt Alignment (60.4\% vs. 27.9\%), also such pattern were observed in Overall Quality (60.4\% vs. 27.9\%).

Overall, these findings conclusively affirm that AMSF’s similarity-aware fusion mechanism generates hybrid images that users perceive as significantly higher in quality, better aligned with the original style references, and more faithful to the provided textual prompts compared to other state-of-the-art training-free methods.

\begingroup
\setlength{\abovecaptionskip}{3pt}   
\setlength{\belowcaptionskip}{0pt}   
\begin{figure}[t]
  \centering
  \includegraphics[width=\linewidth,height=0.30\textheight,keepaspectratio]{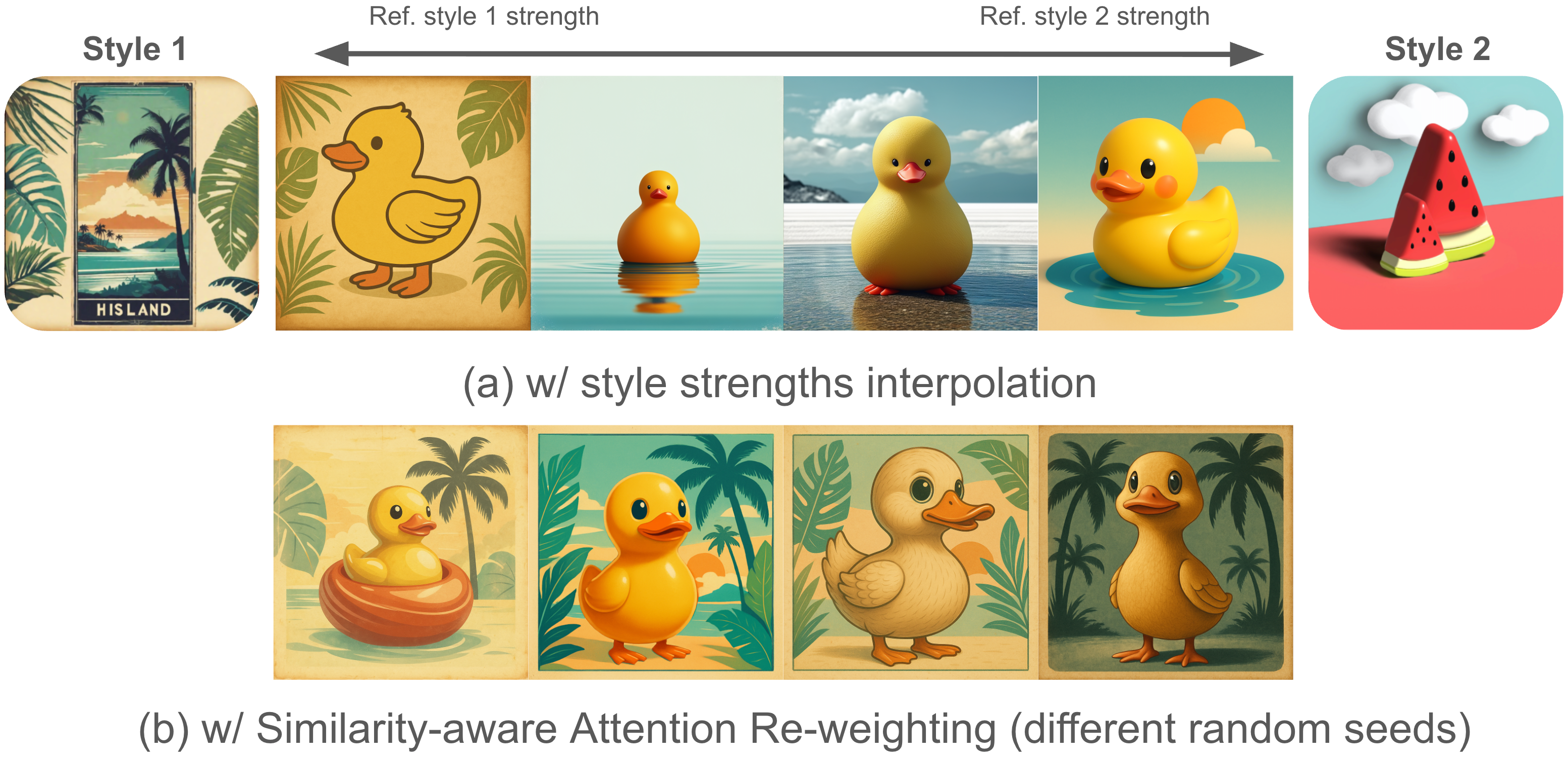}
  \caption{Similarity-aware Attention Re-weighting ablation.
  (a) Linear transition from Style~1 to Style~2 with different style-strength interpolations. 
  (b) Results of SAR with different random seeds, which automatically adjusts the fusion weights instead of using manually tuned strengths.}
  \label{fig:auto_gamma}
\end{figure}
\endgroup

\subsection{Ablation studies}
\label{sec:ablation}

\noindent\textbf{Impact of prompt decomposition.}
One of the most important problems in multi-style fusion is to balance the appearance of different styles as well as the subject or content in the image. Initially, we try to answer a question:
Why don’t we just concatenate the two descriptions,
“\textit{golden-retriever dog in mosaic art style}”
and “\textit{golden-retriever dog in vintage travel poster style}”, and feed the whole sentence to the model?
As shown in Fig.~\ref{fig:prompt_fail}(a),
the cross-attention module weights its attention in proportion to token appearance. For example, if the sentences are merged, the phrase
\textit{golden-retriever dog} appears \textbf{twice}, whereas each style phrase
(\textit{mosaic art style}, \textit{vintage travel poster style}) appears only once.
The duplicated subject tokens therefore dominate the attention map and the
generator reproduces the dog’s anatomy while largely ignoring stylistic cues.

Motivated by this, our semantic token decomposition module gives every semantic component an equal attribution (Fig. ~\ref{fig:msh_cf_overview}).

Each component is encoded once, injected through its own cross-attention layers.
Because the cumulative attention weights is now balanced, the decoder simultaneously preserves the dog’s identity and blends both reference styles,
yielding visible mosaic tessellation and 3D highlights as shown in Fig.~\ref{fig:prompt_fail}(b).

\begingroup
\setlength{\abovecaptionskip}{4pt}
\setlength{\belowcaptionskip}{0pt}
\setlength{\tabcolsep}{2pt} 

\begin{figure}[t]
  \centering
  \begin{tabular}{@{}cccc@{}}
    \subfloat[Style1: pop art style\label{fig:style1}]{
      \includegraphics[width=0.21\textwidth]{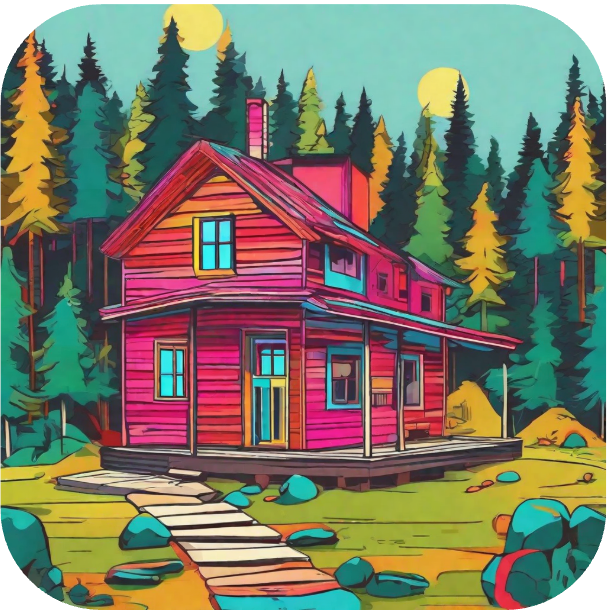}
    } &
    \subfloat[Style2: mosaic art style\label{fig:style2}]{
      \includegraphics[width=0.21\textwidth]{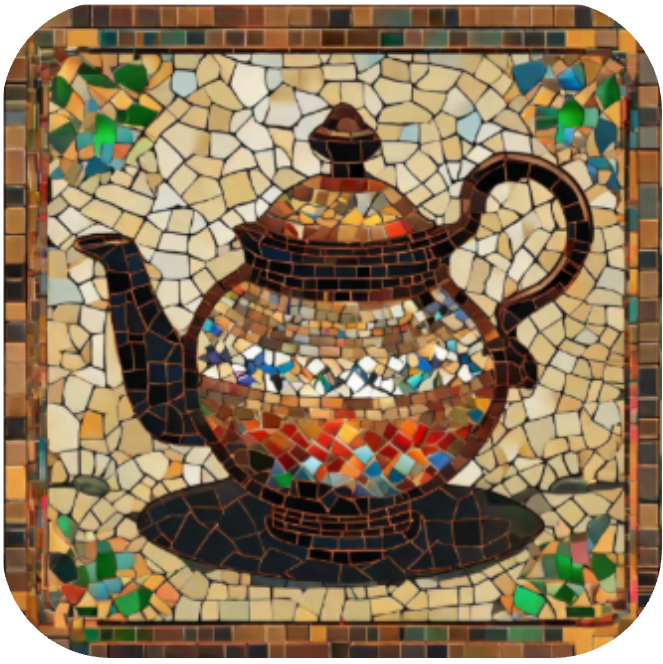}
    } &
    \subfloat[Style3: sticker style\label{fig:style3}]{
      \includegraphics[width=0.21\textwidth]{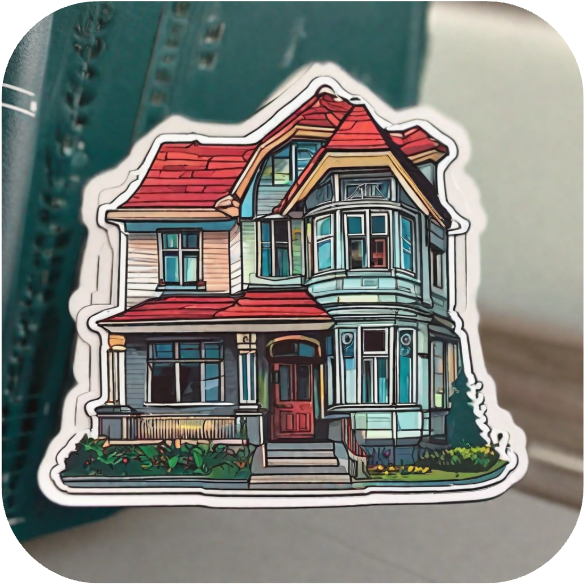}
    } &
    \subfloat[result\label{fig:fused}]{
      \includegraphics[width=0.21\textwidth]{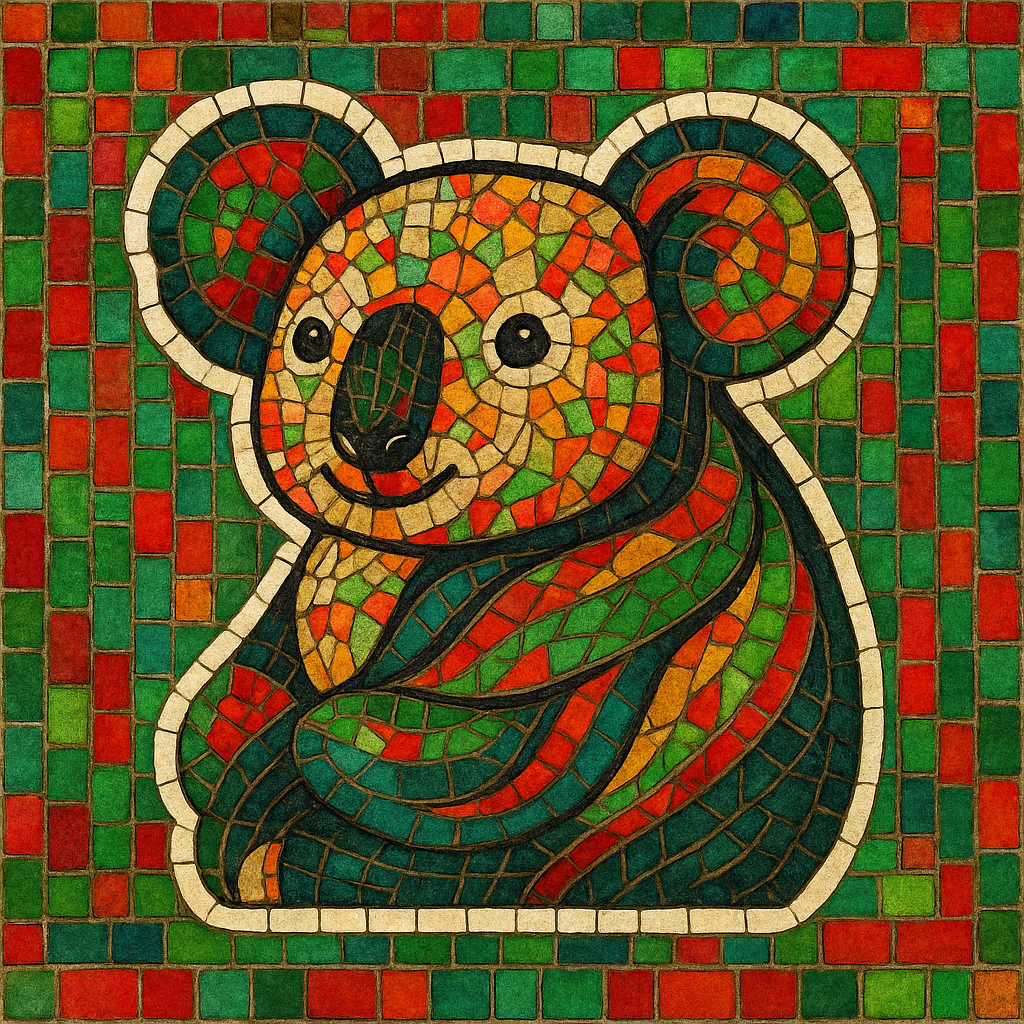}
    }
  \end{tabular}
  \caption{Three‐style composition.
           Our model harmonizes all three references, yielding a hybrid
           aesthetic unattainable with pairwise‐only systems.}
  \label{fig:three_style}
\end{figure}
\endgroup

\noindent\textbf{Effectiveness of Similarity-aware Attention Re-weighting (SAR).}
Considering two style fusion, a naive baseline keeps the two styles and subject weights fixed at
$w_{1}=w_{2}=w_{subject}=0.33$. The users can tune the weights for different styles if the output is less satisfying (Figure~\ref{fig:auto_gamma} (a)).
Such manual balancing ignores the latent-level similarity between the current representation and each reference, while being inefficient.

Our SAR module monitors the global and token-wise similarity gap and raises an exponent to penalize the stronger style’s norm (eq. \eqref{eq:autogamma}).  
This adaptive penalty reduces the over-represented style’s attention on
the fly, restoring a balanced fusion without manual tuning, as shown in Figure~\ref{fig:auto_gamma} (b).

\noindent\textbf{Three-style composition.}
Thanks to the semantic token decomposition and the similarity–aware re-weighting mechanism, our architecture scales beyond the two-style setting without any structural change or retraining.
In Figure~\ref{fig:three_style}, we demonstrate AMSF's fusion results using three different reference styles.
Style1 provides the overall bold color palette and outlines, setting a vibrant base tone.  
Style2 contributes the intricate tile textures and grid-like spatial fragmentation.
Style3 adds clear edge definitions and flattened shading, enhancing cartoon-ish layered structure.
Our results show the Koala's superior performance in extending to more than two styles, as it successfully preserves the white-edged sticker style (style 3), vibrant colors (style 1), and mosaic style (style 2) while automatically balancing the appearance of different components.

\vspace{0.5em}  
\section{Conclusion}
\label{sec:conclusion}

We presented \textbf{Adaptive Multi-Style Fusion (AMSF)}, a reference-based training-free framework that enables
balanced fusion of multiple visual styles using diffusion models.  By (a) semantically decomposing the inputs and
(b) assigning them dynamic similarity-aware weights at every denoising step.
Our method overcomes common issues of style dominance and imbalanced appearance.
Comprehensive experiments demonstrate that AMSF achieves both better quantitative and qualitative results. 
Future work will focus on extending the adaptive weighting scheme to handle geometric style attributes, accelerating inference for high-resolution outputs, and exploring its integration into video diffusion models~\cite{blattmann2023align,ho2022video} to enable user-controllable multi-style video synthesis.

\bibliography{refs}

\end{document}